\documentclass[xcolor = {usenames,dvipsnames}, lettersize,journal]{IEEEtran}

\IEEEoverridecommandlockouts
\usepackage{times}

\usepackage{multicol}
\usepackage[bookmarks=true]{hyperref}
\usepackage{subfigure}
\usepackage{amsthm}
\usepackage{amsmath}
\usepackage{url}  
\usepackage{wasysym}

\usepackage[T1]{fontenc}    
\usepackage{booktabs}       
\usepackage{amsfonts}       
\usepackage{nicefrac}       
\usepackage{microtype}      

\usepackage{soul}
\usepackage[utf8]{inputenc}
\usepackage{enumitem}
\usepackage{balance}
\usepackage{fontawesome}
\usepackage{multirow,array}
\usepackage{tikz}
\usepackage{pgfplots}
\usetikzlibrary{pgfplots.dateplot}
\usepackage{pgfplotstable}

\usepackage{filecontents}
\usepgfplotslibrary{fillbetween}
\usetikzlibrary{patterns}
\usetikzlibrary{backgrounds}
\usetikzlibrary{arrows,automata}
\graphicspath{ {images/} }
\usepackage{wrapfig}
\usepackage{diagbox}
\usepackage{adjustbox}
\usepackage{tcolorbox}
\newtheorem{proposition}{Proposition}

\begin{document}

\title{Inference of Human's Observation Strategy for Monitoring Robot's Behavior based on a Game-Theoretic Model of Trust}
\author{Zahra Zahedi$^*$, Sailik Sengupta$^*$, and Subbarao Kambhampati
    \thanks{$*$ Z. Zahedi, and S. Sengupta have contributed equally.}
    \thanks{Z. Zahedi and S. Kambhampati are with the School of Computing and AI, Arizona State University (e-mail: \{zzahedi, rao\}@asu.edu)}
    \thanks{S. Sengupta is with \faAmazon WS AI Labs, but the work was done while at Arizona State University (e-mail: sailiks@amazon.com)}
    \thanks{This work has been submitted to the IEEE for possible publication. Copyright may be transferred without notice, after which this version may no longer be accessible.}}

\maketitle


\begin{abstract}

In scenarios where a robot generates and executes a plan, there may be instances where this generated plan is less costly for the robot to execute but unexpected and unsafe from human's point of view. When the human acts as a supervisor and is held accountable for the robot's plan, the human may be at a higher risk if the incomprehensible behavior is deemed to be infeasible or unsafe.
In such cases, the robot, who may be unaware of the human's exact expectations, may choose to execute (1) the most constrained plan (i.e. one preferred by all possible supervisors) incurring the added cost of executing highly sub-optimal behavior when the human is monitoring it and (2) deviate to a more optimal plan when the human looks away. While robots do not have human-like ulterior motives (such as being lazy), such behavior may still occur because the robot has to cater to the needs of different human supervisors.
In such settings, the robot, being a rational agent, may take any chance it gets to deviate to a lower cost plan.
On the other hand, continuous monitoring of the robot's behavior is often difficult for humans because it costs them valuable resources (e.g., time, cognitive overload, etc.). Thus, to optimize the cost of monitoring while ensuring the robots follow the {\em safe} behavior and to assist the human to deal with the possible unsafe robots, we model this problem in a game-theoretic framework of trust. In settings where the human does not initially trust the robot, pure-strategy Nash Equilibrium provides a useful policy for the human. Unfortunately, we show the formulated game often lacks a pure strategy Nash equilibrium. Thus, we define the concept of a trust boundary over the mixed strategy space of the human and show that trust boundary helps in discovering optimal monitoring strategies. We conduct humans subject studies that demonstrate (1) the need for coming up with optimal monitoring strategies, and (2) the benefits of using strategies suggested by our approach.
\end{abstract}

\begin{IEEEkeywords}
Human-Robot interaction, trust boundary, game-theoretic model, safe behavior.
\end{IEEEkeywords}
\IEEEpeerreviewmaketitle
\section{Introduction}
\IEEEPARstart{I}{n} a multi-agent scenario involving a robot ($R$), who is making and executing a plan (or policy) in the world, and a human supervisor ($H$), who monitors the robot's action and is held responsible for $R$'s behavior, the notion of trust becomes key for successful interaction. When the supervisor trusts the robot, they do not need to always spend their valuable resources such as time and cognitive effort in monitoring or intervening in the robot's plan (or execution of these plans). While it is possible to develop trust in longitudinal settings \cite{chen2018planning, xu2015optimo}, in one-off interactions (where no warranted trust exists \cite{jacovi2021formalizing}) conventional wisdom often guides the supervisor to spend all their time monitoring the robot's behavior to ensure that it adheres to their expectations (making it too resource-intensive for the human), especially since previous studies show that negligence at monitoring time affects robot performance \cite{goodrich2003seven}. In this paper, we seek to challenge this idea and show that a human can consider resource-efficient monitoring strategies.

There are cases when a robot's expectation may deviate from its supervisor's expectations. First, a robot may have side-goals that do not align with the supervisor's expectation. For example, an autonomous car ride-sharing (or, in general, robot-as-a-service) may have certain expectations from its supervisor (eg. travel on shortest routes) but may need to adhere to passenger's expectation (eg. avoid hilly roads) that are in conflict with one another. Second, the worker robot may not be aware of the human's exact model $M^{R}_{H}$ that describes the safety requirements the supervisor has in mind. 
Hence, when the human does not observe the robot's plan or its execution, the robot may choose to execute a less costly plan that is deemed unsafe (by the human). To handle such scenarios, we formally model the inference problem related to the finding a monitoring strategy for the human supervisor that saves their valuable resources (time, cognitive overload) while ensuring that the robot sticks to the expected behavior and achieves the goal.

Specifically, we introduce a notion of trust that a human supervisor $H$ places on a worker robot $R$ when $H$ chooses to {\em not observe} $R$'s plan or its execution, by modeling the interaction in a game-theoretic framework of trust motivated by \cite{sankaranarayanan2007towards}. In our case, the robot is unaware of the human's exact model $M^{R}_{H}$, but has knowledge about all the possible sets $\mathcal{M}^{R}_{H}$ of safety constraints the human might have, i.e., $M^{R}_{H} \in \mathcal{M}^{R}_{H}$. This uncertainty about the human's model that $R$ has can be reflected in the utilities of the players, making our formulated game a Bayesian one. Without prior interaction (and thus, a lack of trust) if $H$ does not observe $R$, $R$ will always deviate to a plan that is less costly for itself. In this section, we show that $H$ can devise a probabilistic observation strategy that ensures that (1) $R$ does not deviate from executing the safest plan (i.e., executable in all the models of $\mathcal{M}_{H}^{R}$) and also, (2) $H$ saves valuable resources (such as time, effort, etc.) as opposed to continually monitoring $R$.
    
In addition to providing a novel type of service that can assist $H$ on when to supervise $R$ to ensure expected behavior, we also explore if such a service is useful in practice by performing human studies and to figure out what are the natural strategies they would follow. First, we show that in such supervision or monitoring scenarios, humans may either be risk-averse (ensuring that the robot does the right thing, no matter the monitoring cost) or risk-taking (in the hope to minimize their cost, will choose to cut down their monitoring time). These results justify the Bayesian modeling of our human player in the game-theoretic framework for the supervision scenario. Second, we show, in contrast to work in existing human-aware planning scenarios where humans are asked to monitor the robot all the time \cite{kulkarni2016explicable,dragan2013legibility}, humans often deviate to more split-time strategies where some of the time, that is originally meant for monitoring, can be used for other tasks and still ensure that the robot adheres to constraints.
Thus, it makes sense to analyse the supervision scenario formally and provide human agents with optimal monitoring strategies that let them maximize their utility while ensuring the supervised agent $R$ does not execute behavior that is either unsafe or fails to achieve the goal. Lastly, we conduct another human study when the optimal strategy is suggested to the human, and demonstrate that suggesting the optimal strategy as computed by our approach will help the human to come up with better monitoring strategy. The paper concludes with some direction for future extension.
\section{Related Work}
Our work is situated in the middle of the spectrum that ranges from fully cooperative settings to fully-adversarial ones. In fully-cooperative settings, the robot only considers the human's goals and thus, can only exhibit undesirable behavior because of either impreciseness in or differences between its own model $M^R$ and the human's expectation $M^{R}_{H}$. 

In motion and task planning, researchers argue that if the robot follows a plan that adheres to the human's expectation, i.e., is optimal in $M_{H}^{R}$; then these plans are deemed to be explicable~\cite{zhang2017plan}, legible~\cite{dragan2013legibility}, or adhers to social norms~\cite{kockemann2014grandpa}. They assume that the need for $R$ to be explicable, legible, etc. is because the human is continuously observing or monitoring the robot. Although they do not explicitly discuss, in scenarios where the human is not observing the robot, it may deviate to a plan that is optimal in $M^R$. In our setting, this deviation can result in the violation of safety constraints and hence we want to ensure that even when the human is not spending all their resources in observing $R$, the robot does not deviate from the safe plan $\pi_s$.
Furthermore, the existing works \cite{zhang2017plan,dragan2013legibility,kockemann2014grandpa} assume that all the humans who observe the robot have the same expectation, i.e., $\mathcal{M}_{H}^{R}$ is a singleton set, which is either fully known beforehand or can be easily learned. Some recent works, such as \cite{hadfield2016off}, that try to address this concern, consider the imprecise specification of the human's reward (which can be a part of $M_{H}^{R}$). Then they show how it results in the robot executing undesired behaviors that may be deemed unsafe. Eventually, they conclude that some uncertainty about $M_{H}^{R}$ may result in $R$ doubting its current behavior as unsafe and in turn, letting the human take control (switch it off) if necessary.
Unfortunately, they consider that $R$'s objective is solely to maximize the human's reward and thus, robots have no reason to think of other rewards. Although the robot may not have ulterior motives like human agents, the assumption falls flat when the robot is (1) rented out as a service by a third-party agent for helping a particular human (autonomous car offered by ride-sharing apps), or (2) is catering to the needs of multiple supervisors. In such scenarios, a single human's reward is not its sole reward anymore. We seek to address such scenarios in this work. Although, similar to our work, researchers have looked at the idea of considering multiple human models, they mostly address the problem generating robust explanations \cite{sreedharan2017explanations}. Their produced explanation is to soothe the human; neither can guarantee behavior produced by $R$ adheres to human's expectation. Other methods where the supervisor communicates implicit constraints \cite{johnson2020impact}, or their preferences \cite{kim2017collaborative} may not work in our scenario, as a two-way channel is necessary for the robot to identify conflicting constraints, communicate back to the supervisor and convince $H$ the rational behind their behavior.

Given that we are trying to find a monitoring strategy for the human supervisor so that the robot always chooses to execute $\pi_s$ even if there exists uncertainty about the human's model, we should also consider works in the other end of the spectrum that deal with adversarial monitoring in physical \cite{paruchuri2008playing,sinha2015physical} and cyber domains \cite{sengupta2017game,schlenker2018deceiving}. A key difference with these works is that they lack any notion of cooperation. In our case, if the robot $R$ is unable to achieve the (team) goal due to violation of certain constraints and insufficient monitoring, it results in an inconvenience for $H$ too, who will then be held responsible for their failure to (1) ensure safety or (2) achieve the goal. Beyond these, our framework should be seen as a first-step towards repeated game modeling that will allow us to consider the development of trust on the robots and eventually, finding methods to incentive the robot to identify and respect that trust. Such intentions are clearly missing in adversarial settings. Lastly, the notion of mixed strategies that are used in most of these works does not sit well with our scenario because the probabilistic guarantees about the robot behaving safely might not be an acceptable solution in our settings. Thus, we can conclude that although our problem shares properties of both fully cooperative and fully adversarial settings, it exhibits significant differences to reside in the middle of the aforementioned spectrum.
\section{Game Theoretic Formulation}
	Before describing the game-theoretic formulation--the actions and the utilities of the agents-- we first clearly highlight the assumptions made about the two agents.
	\subsection{Assumptions about the Agents}
	
	\subsubsection*{\textbf{The human $H$}} who is a supervisor in our setting, has the following characteristics:
	\begin{enumerate}
	    \item $H$ has a particular model of the robot $R$, denoted as $M_{H}^{R}$ that belongs to some set of possible models $\mathcal{M}_{H}^{R}$.
        \item Upon observation of the plan that $R$ comes up with or its execution, if $H$ believes the plan is risky (i.e., is inexecutable or unsafe in their model $M_{H}^{R}$ of the robot), $H$ can stop the execution at any point in time. If $H$ stops the robot $R$ from executing its plan, $H$ incurs some cost of inconvenience for not having achieved the team goal $G$ or because $H$ should stop the robot and make the robot to do the safe plan. This seems pragmatic because $H$, being the supervisor, will be held responsible for it.
        \item $H$ has a positive cost for observing the robot's plan or the plan's execution.
	\end{enumerate}
	\subsubsection*{\textbf{The Robot $R$}} who is the agent being monitored, has the following capabilities and assumptions associated with it:
    \begin{enumerate}
        \item $R$ is uncertain about the human's model of it, i.e., $M_{H}^{R}$, but knows that it belongs in the set of possible models $\mathcal{M}_{H}^{R}$.
        \item $R$, given a sequential decision making problem, can come up with two plans-- (1) a safe plan ($\pi_s$) that is executable in all models $\in \mathcal{M}_{H}^{R}$ and (2) a risky plan ($\pi_{pr}$) that is executable in a subset of $\mathcal{M}_{H}^{R}$ but in-executable (or unsafe) in the other models.
        \item There are costs for coming up with the plans $\pi_s$ and $\pi_{pr}$ and executing them. Also, since $R$ may have to work on other goals or cater to the needs of other supervisors, it would like to execute $\pi_{pr}$ if it can get away with it.
        \item It incurs a cost for not achieving the team's goal $G$. This happens when the human observes the plan or execution and stops it midway (due to safety concerns).
        \item The robot is not malicious and thus, does not lie. It won't bait-and-switch by showing one plan to $H$ (that looks safe) and then executing another.
    \end{enumerate}
    
   With these assumptions in place, we can now define each players' pure strategies and their utility values which will encode the uncertainty about the types of human supervisor, turning the game a Bayesian one.

    \begin{table*}[t!]
	\begin{centering}
    \begin{adjustbox}{width=.95\textwidth}
		\begin{tabular}{ |c||c|c|c| } 
			\hline
			 & $O_{P,\neg E}$ & $O_{\neg P, E}$ & NO-OB  \\
			\hline
			\hline
			$\pi_{pr}$
            & \shortstack { 
              ${\color{MidnightBlue}-C^{H}_{P}(\pi_{pr})-I_P^{H}(\pi_{pr})}$, \\
              ${-C^{R}_P(\pi_{pr})-C_{\tilde{E}}^R(\pi_{pr})}-C_{\tilde{G}}^R$
            }
            & \shortstack {
              ${\color{MidnightBlue}-C^{H}_{E}(\tilde{\pi}_{pr})-I_E^{H}(\hat{\pi}_{pr})}$, \\
              ${-C^{R}_P(\pi_{pr})-C_E^R(\tilde{\pi}_{pr})-C_{\tilde{G}}^R}$
            }
            & \shortstack {
              ${\color{MidnightBlue}-V^{H}_{I}(\pi_{pr})}$, \\    
              ${-C^{R}_P(\pi_{pr})-C_E^R(\pi_{pr})}$
        	} \\
			\hline
			$\pi_s$
            & \shortstack {
              ${\color{MidnightBlue}-C^{H}_{P}(\pi_s)\overbrace{-I_P^{H}(\pi_s)}^{0}}$, \\
              ${-C^{R}_P(\pi_s)-C_E^R(\pi_s)}$ 
            }
            & \shortstack {
              ${\color{MidnightBlue}-C^{H}_{E}(\pi_s)\overbrace{-I_E^{H}(\hat{\pi}_s)}^{0}}$, \\
              ${-C^{R}_P(\pi_s)-C_E^R(\pi_s)}$
            }
            & \shortstack {
              ${\color{MidnightBlue}\overbrace{-V^{H}_{I}(\pi_s)}^{0}}$, \\
              ${-C^{R}_P(\pi_s)-C_E^R(\pi_s)}$
            } \\
			\hline
		\end{tabular}
        \end{adjustbox}
        \vspace{3pt}
        \caption{Normal-form game matrix for modeling the robot-monitoring scenario. $R$ ($H$) is the row (column) player.}
        \label{table:normal-game}
	\end{centering}
\end{table*}
	\subsection{Player Actions}
	In the normal form game matrix shown in \autoref{table:normal-game}, the row-player is the robot $R$ who has two pure strategies to choose from-- the plans $\pi_{pr}$ and $\pi_s$ (as described above). The column player is the human $H$ who has three strategies-- (1) to only observe the plan made by the robot $O_{P,\neg E}$ and decide whether to let it execute (or not), (2) to only observe the execution $O_{\neg P,E}$ and stop $R$ from executing at any point, and (3) not to monitor (or observe) the robot at all (NO-OB).
    
    A few underlying assumptions that are inherent part in our action definitions are (1) the robot cannot switch from a plan (or a policy) it has committed to a different one in the execution phase and (2) the human only stops the robot from executing the plan if they believe that the robot's plan does not achieve the goal $G$ as per their actual model, i.e. the robot's plan is deemed in-executable (or unsafe) given the domain model $M_{H}^{R}$.
	\subsection{Utilities}
    The utility values for both the players are indicated in the game-matrix shown in \autoref{table:normal-game}. In each cell, corresponding to the pure-strategy pair played by the two players, the numbers shown at the bottom in black are the utility values for $R$ while the ones at the top in {\color{MidnightBlue}blue} are the utility values for $H$. We now describe the utilities for each player in our formulated game and later, in the experimental section, talk about how they can be obtained in the context of existing task-planning domains.
    
    \subsubsection*{\textbf{Robot's Utility Values}}
    
    We first describe the notation pertaining to the robot utilities and then use them to compose the utilities for each action pair.\\[-2em]
	\begin{tcolorbox}
		 $C^{R}_{P}(\pi)$ -- Cost of making a plan $\pi$.\\[3pt]
         $C^{R}_{E}(\pi)$ -- Cost to robot for executing plan $\pi$.\\[3pt]
         $C_{\tilde{G}}^{R}$ -- Penalty of not achieving the goal.
	\end{tcolorbox}%

    Note that we use the variables $C$ to represent a non-negative cost or penalty. Thus, the rewards for the robot $R$ shown in \autoref{table:normal-game} have a negative sign before the cost and penalty terms. As the human may choose to stop the execution of a plan midway, the robot might have executed a part of the original plan. We denote this partial plan by $\hat{\pi}_{pr}$. Given this, the term $C^{R}_{E}(\hat{\pi}_{pr})$ represents the cost of executing the partial plan.\footnote{Depending on where the human will stop the robot, the cost for the partial plan is different.}
    
    The uncertainty in the robot's mind as to whether a particular supervisor type will let it execute the plan $\pi_{pr}$ to completion can now be captured using the variable $C_{\tilde{G}}^R$ that represents the cost of not achieving the goal.
    Before we discuss how one can model the variable $C_{\tilde{G}}^R$, let us first briefly talk about the robustness $r$ of the plan $\pi_{pr}$. The parameter $r \in (0,1]$ represents the fraction of models in $\mathcal{M}_{H}^{R}$ where the plan $\pi_{pr}$ is executable (and thus, safe). A way of obtaining this value for deterministic planning problems could be the use of model counting \cite{nguyen2017robust}.
    For a given $r$, an idea to model the cost associated with not achieving the goal is to consider $C_{\tilde{G}}^R$ as a random variable drawn from the Bernoulli distribution s.t. $C_{\tilde{G}}^R$ is a non-zero penalty if the plan is not robust enough for a given human (with probability $1-r$) or zero if it is (with probability $r$).
    
    Whenever the cost of not achieving the goal is equal to zero, it means that the robot's plan $\pi_{pr}$ (or its execution) was observed by $H$ and not stopped by them. If the human chooses to observe the plan before execution, then the cost incurred by the robot for executing the plan $\pi_{pr}$ can be represented as,
    \begin{eqnarray}
        C_{\tilde{E}}^R(\pi_{pr}) =  
            \begin{cases}
                C_E^R(\pi_{pr}) &\quad\text{if } C_{\tilde{G}}^R=0\\
                0 &\quad\text{\textit{o.w.}}
            \end{cases}
    \end{eqnarray}%
    If the supervisor $H$, on the other hand, chooses to monitor the execution directly, then the cost of execution would be,
    \begin{eqnarray}   
        C_E^i(\tilde{\pi}_{pr}) = 
            \begin{cases}
                C_E^i(\pi_{pr}) &\quad\text{if } C_{\tilde{G}}^i=0 \hspace{10pt} i\in \{R, H\}\\
                C_E^i(\hat{\pi}_{pr}) &\quad\text{\textit{o.w.}}
            \end{cases}
    \end{eqnarray}%
    In the formulated game, the robot {\em has to} come up with a plan (even though it may not be allowed to execute it). Thus, the cost to come up with a  plan ($\pi_s$ or $\pi_{pr}$) has to be considered for all the utility values (in the respective rows). In the case of $\pi_s$, since it is executable in all the models of $\mathcal{M}_{H}^{R}$, there is no chance that $H$ will stop its execution and thus, no chance of incurring a penalty for not achieving the goal.
    
    Note that the cost of executing a plan that adheres to all the models in $\mathcal{M}_{H}^{R}$ is going to be high because it respects all the constraints enforced by all the model (corresponding to all possible humans). On the other hand, executing a plan $\pi_{pr}$ that respects constraints corresponding to a subset of models in $\mathcal{M}_{H}^{R}$ would be less costly to execute. Thus, it is natural to assume $C^{R}_{E}(\pi_{pr}) \leq C^{R}_{E}(\pi_s)$.
    
    Similarly, coming up with $\pi_{pr}$ may often be easy if the value of $r$ is small while coming up with the plan $\pi_s$ that is guaranteed to work in all the models of $\mathcal{M}_{H}^{R}$ may take a considerable longer amount of time. Hence, even for the planning time, we make the logical assumption that $C^{R}_{P}(\pi_{pr}) \leq C^{R}_{P}(\pi_s)$.

	\subsubsection*{\textbf{Human's Utility Values}}
    
    We first describe the notations and then use them to obtain the various utilities for the human.\\
    	\begin{tcolorbox}
    	$C^{H}_{P}(\pi)$ -- Cost w.r.t. human's resources of observing the plan $\pi$ made by the robot.\\[3pt]
        $C^{H}_{E}(\pi)$ -- Cost w.r.t. human's resources of observing the robot execute the plan $\pi$.\\[3pt]
        $V^{H}_{I}(\pi)$ -- Cost incurred by the human, who was responsible for the robot's plan for violating a constraint that it had set for the robot to follow and being ignorant about it. Note that $V^{H}_{I}(\pi_s)=0$\\[3pt]
        $I_P^{H}(\pi)$ -- Inconvenience to the human if they see a plan that it cannot let the robot execute. Note that $I_P^{H}(\pi_s)=0$.\\[3pt]
        $I_E^{H}(\pi)$ -- Inconvenience to the human if the human observes the execution of an unsafe plan and it has to intervene or stop from execution. Note that $I_E^{H}(\pi_s)=0$.
    	\end{tcolorbox}
    

    \noindent Note that, in our setting, the human supervisor $H$ will be held responsible for not achieving the goal. This happens when $H$ has to stop the robot from executing the plan $\pi_{pr}$. The inconvenience cost can be represented using a negative utility for the human and is denoted using the last two notations.
    
    In our setting, after the robot comes with a plan, unless it is $\pi_s$, the human $H$ is not sure if the robot's strategy will be executable (or safe) in their model $M_{H}^{R}$ because the plan $\pi_{pr}$ is executable in a subset of models which may not contain H's model $M_{H}^{R}$. Thus, they have some uncertainty over the variables $V^{H}_{I}(\pi)$, $I_P^{H}(\pi)$ and $I_E^{H}(\pi)$. Thus, similar to the robots penalty, they can be represented as random variables sampled from a Bernoulli distribution.
    
    With probability $(1-r)$, when the robot chooses to come up (and then execute) the plan $\pi_{pr}$, if the human does not observe either of the two processes, i.e., chooses NO-OB, then it is natural to assume that the human, who is going to be held responsible for the plan will eventually find out that constraints set by them was violated. The cost incurred by the supervisor in this case (i.e. $R$ plays $\pi_{pr}$ and $H$ plays NO-OB), should be the highest because (1) the robot, without $H$'s knowledge, violated some safety or social norm (that was necessary for a plan to achieve the goal in $M_{H}^{R}$), (2) $H$ will be held accountable for it, and (3) blamed for not fulfilling their supervisory duties. Thus, we have,
    \begin{eqnarray}
      \label{eqn:v_great_1}
      V^{H}_{I}(\pi_{pr}) &>& C^{H}_{P}(\pi_{pr})+I_P^{H}(\pi_{pr})\\
      \label{eqn:v_great_2}
      V^{H}_{I}(\pi_{pr}) &>& C^{H}_{E}(\tilde{\pi}_{pr})+I_E^{H}(\hat{\pi}_{pr})
    \end{eqnarray}
    We also consider the cost of observing the execution of a plan is greater than cost of observing the plan, i.e.%
    \begin{eqnarray}
      C^{H}_{E}(\pi) > C^{H}_{P}(\pi)
    \end{eqnarray}%
    and the inconvenience caused by execution of a probably risky (partial) plan is greater than inconvenience cause by just observing the plan because no damage has yet been done. Thus,
    \begin{eqnarray}
    \label{eqn:I_great_I}
      I_E^{H}(\hat{\pi}_{pr}) > I_P^{H}(\pi_{pr})
    \end{eqnarray}%
    Lastly, note that when the robot comes up with a plan $\pi_s$ that is executable in all the models of $\mathcal{M}^{R}_{H}$, the inconvenience ($I^{H}_{P}(\pi_s)$ and $I^{H}_{E}(\pi_s)$) and responsibility ($V_{I}^{H}(\pi_s)$) costs are zero. This is indicated used curly braces in \autoref{table:normal-game}.


\section{Game-Theoretic Notion of Trust}
    In this section, we first define a notion of trust in the formulated game shown in \autoref{table:normal-game}. $H$ has three actions and as one goes from left to right, the amount of trust $H$ places in $R$, as defined in \cite{sankaranarayanan2007towards}, increases. Consider the human chooses not to observe the robots plan or its execution, i.e., chooses NO-OB. Clearly, $H$ exposes itself a vulnerability because if $R$ comes up with and executes $\pi_{pr}$, it can result in $H$ getting a high negative reward $V_{I}^{H}$. On the other hand, the robot may choose to respect the human's trust by selecting $\pi_s$ and therefore, not exploit the vulnerability that presents itself when the human plays No-OB. On the other hand, if the human chooses to observe the plan ($O_{P,\neg E}$), the human is exposed to the least amount of risk because the robot plan, even before it can execute the first action, is verified by the human.
    
    Note that $H$ incurs a non-negative cost when playing the action $O_{P,\neg E}$ because it has to spend both time and effort in observing the robots plan and then deciding whether to let it execute. In scenarios when $H$ cannot fully trust the robot and they have to play $O_{P,\neg E}$ or $O_{\neg P, E}$, they will incur the cost of constant monitoring. We now discuss this case of {\em no-trust} in our game and see if it possible to minimize this cost.

\begin{figure*}[t]
\centering
\begin{adjustbox}{width=0.45\textwidth}
\begin{tikzpicture}[
            >=stealth',
            shorten >=4pt,
            auto,
            node distance=3cm,
            state/.style={circle,inner sep=2pt}
            ]
            \centering
            \node[state] (Bot) [text width=1cm, align=center] {{\Huge \faAndroid}\\{\tiny Robot}};
            \node[state] (Coffee)  [above of=Bot, text width=1cm, align=center] {{\LARGE \faCoffee}\\{\tiny Kitchen}};
            \node[state] (Box) [left of=Bot, text width=1cm, align=center] {{\LARGE \faDropbox}\\{\tiny Reception}};
            \node[state] (Human) [right of=Bot, text width=1cm, align=center] {{\Huge \faFemale}\\{\tiny Employee}};
            \node[state] (Eye) [above of=Human, text width=1cm, align=center] {\LARGE \faEye};
            
            \path (Bot)
                edge  [->, ForestGreen!80, dashed, bend left=20]  node {$1$} (Coffee)
                edge  [->, ForestGreen!80, dashed, bend left=10]  node {$3$} (Human)
                edge  [->, ForestGreen!80, dashed, bend left=20]  node {$5$} (Box)
                edge  [->, ForestGreen, dashed, bend left=40]  node {$7$} (Human);
            \path (Coffee)
                edge  [->, ForestGreen!80, dashed, bend left=20]  node {$2$} (Bot);
            \path (Human)
                edge  [->, ForestGreen!80, dashed, bend left=10]  node {$4$} (Bot);
            \path (Box)
                edge  [->, ForestGreen!80, dashed, bend left=20]  node {$6$} (Bot);
            
            \draw[dashed] (1,2) -- (Eye);
            \draw[dashed] (1.5,1.5) -- (Eye);
            \draw[dashed] (2,1) -- (Eye);
\end{tikzpicture}
\end{adjustbox}
\quad
\begin{adjustbox}{width=0.47\textwidth}
\begin{tikzpicture}[
            >=stealth',
            shorten >=4pt,
            auto,
            node distance=3cm,
            state/.style={circle,inner sep=2pt}
            ]
            \centering
            \node[state] (Bot) [text width=1cm, align=center] {{\Huge \faAndroid}\\{\tiny Robot}};
            \node[state] (Coffee)  [above of=Bot, text width=1cm, align=center] {{\LARGE \faCoffee}\\{\tiny Kitchen}};
            \node[state] (Box) [left of=Bot, text width=1cm, align=center] {{\LARGE \faDropbox}\\{\tiny Reception}};
            \node[state] (Human) [right of=Bot, text width=1cm, align=center] {{\Huge \faFemale}\\{\tiny Employee}};
            \node[state] (Eye) [above of=Human, text width=1cm, align=center] {\LARGE \faLowVision};
            
            \path (Bot)
                edge  [->, BrickRed!80, dashed, bend left=20]  node {$1$} (Coffee)
                edge  [->, BrickRed!80, dashed, bend left=10]  node {$5$} (Human)
                edge  [->, BrickRed!80, dashed, bend left=20]  node {$3$} (Box);
            \path (Coffee)
                edge  [->, BrickRed!80, dashed, bend left=20]  node {$2$} (Bot);
            \path (Box)
                edge  [->, BrickRed!80, dashed, bend left=20]  node {$4$} (Bot);
            
            \draw[dotted] (1,2) -- (Eye);
            \draw[dotted] (1.5,1.5) -- (Eye);
            \draw[dotted] (2,1) -- (Eye);
\end{tikzpicture}
\end{adjustbox}
\caption{The two plans, i.e the safe plan {\color{ForestGreen!80} $\pi_s$} (left) and the probably-risky plan {\color{BrickRed!80} $\pi_{pr}$} (right) for the robot-delivery scenario.}
\label{fig:scenarios}
\end{figure*}

\subsection{The No-Trust Scenario}

In this setting, $H$ should never play an action that exposes them to a risk of a high negative utility because it does not trust $R$ (who will play $\pi_{pr}$ if $H$ plays NO-OB). In such scenarios, if there exists a pure-strategy Nash Equilibrium, then the players should play it because neither of the players can deviate to get a better utility \cite{sankaranarayanan2007towards}. In our setting, this depends on the value of $r$, if $r$ is high and close to $1$, it means that for most of the models $M_H^R \in$ $\mathcal{M}_H^R$, the plan $\pi_{pr}$ is executable. Given we consider a Bayesian game, in order to have the Nash Equilibrium we should satisfy the following condition over the expected utility,
\begin{eqnarray}
\label{nash condition}
(1-r)V_I^H(\pi_{pr}) < C_P^H(\pi_{pr})+(1-r)I_P^H(\pi_{pr}) \\
C_P^R(\pi_{pr})+(1-r)C_{\tilde{G}}^R+rC_E^R(\pi_{pr}) < C_P^R(\pi_s)+C_E^R(\pi_s)
\end{eqnarray}
As $r \rightarrow 1$, we can guarantee that $(\pi_{pr}, NO-OB)$ is the Nash equilibrium because $\pi_{pr}$ is executable in a large majority of the models in $\mathcal{M}_H^R$. In this case, with high probability, the human observer (whose model ${M}_H^R$ is sampled from the set $\mathcal{M}_H^R$) has no preference about the robot using $\pi_s$ over $\pi_{pr}$. Thus, with high probability, they will not incur $V^{H}_{I}$. Therefore, it makes sense for the robot $R$ to choose $\pi_{pr}$ that is less costly.

Note that the above scenario is where $r$ is closer to $1$ is highly unrealistic. It can only occur in domains where executing $\pi_{pr}$ does not result in catastrophic circumstances or lead to in-feasibility, implying the distinction between $\pi_{s}$ and $\pi_{pr}$ is hardly present. In most real world settings, this would hardly be the case (i.e. $r$ will be much lower than $1$), leading to the following proposition.


\begin{proposition}
The game defined in \autoref{table:normal-game} has no pure strategy Nash Equilibrium where $\pi_{pr}$ is not executable in some of the models in the set $\mathcal{M}^R_H$.
\end{proposition}
\noindent {\em Proof.} The formulated game in this paper is a Bayesian game with two player types for the human. The first type is the one where $\pi_{pr}$ is executable in the model $M_H^R$ in $\mathcal{M}_H^R$, so $C_{\tilde{G}}^R = I_P^H(\pi_{pr})=I_E^H(\hat{\pi}_{pr})= V_I^H=0$, and the second type is represents the set of humans whose models are in $\mathcal{M}_H^R$ and $\pi_{pr}$ is not executable in them. Consequently, $C_{\tilde{G}}^R$, $I_P^H(\pi_{pr})$, $I_E^H(\hat{\pi}_{pr})$ and $V_I^H \not= 0$. Given a pure strategy Nash Eq. (as per \autoref{nash condition} and 8) only exists for the former, this game has no pure strategy Nash Equilibrium in the second case (with probability of $1-r$, as $r$ is also the probability of former type).
\hfill \qedsymbol

\subsubsection*{\textbf{Absence of Pure Strategy Nash Equilibrium}}

The absence of a pure-strategy Nash eq. makes it difficult to define a human's best course of action in the no-trust setting \cite{sankaranarayanan2007towards}. Furthermore, existing works that assume the human should always monitor the robot's plan or behavior to ensure the robot plan is explicable \cite{zhang2017plan} or legible \cite{dragan2013legibility} (similar to $\pi_s$ in our setting) fail to account for the human's monitoring. This is unrealistic (rather, too costly) for $H$ to always select $O_{P,\neg E}$ or $O_{\neg E, P}$ in real-world settings. Furthermore, the notion of a mixed-strategy (Nash) equilibrium is inappropriate in our setting because a probabilistic play by $R$, i.e. choosing a risky plan with some non-zero probability cannot guarantee safety or feasibility for all human supervisors. Thus, we devise the notion of a trust boundary that allows the human to play a mixed strategy that reduces their cost of monitoring but ensures the robot always sticks to selecting (and executing) $\pi_{s}$.

\subsubsection*{\textbf{Trust Boundary}}
Consider a human chooses the mixed strategy $\vec{q} = [(1-q_E-q_N), q_E, q_N)]^T$ over the actions $O_{P, \neg E}, O_{\neg P, E}$ and NO-OB respectively. First, let us discuss what it means intuitively if all the values are non-zero. The human probabilistically chooses to look into the plan or execution of a plan done by the robot they are supervising. In many human-human scenarios, such uncertainty (eg. parents may come back) on the part of the supervising agent (say, parents) might instill a fear in the supervised agent (say, children) of getting caught if the latter choose to betray the supervisor (say, watching TV ($\pi_{pr}$) instead of studying ($\pi_s$) when the parents are out). Note that a strategy in  $q_N=1$ will always result in the robot choosing the probably risky plan (especially in our single-step game). Thus, in order to ensure that the robot cannot deviate away from the making and executing $\pi_s$, we have to ensure that the expected utility ($U$) for the robot given $\vec{q}$ is greater for $\pi_s$ than for $\pi_{pr}$. Using the values defined in \autoref{table:normal-game}, this can be formally stated as follows.
{\small
\begin{eqnarray}
\label{eqn:not}
\mathbb{E}_{\vec{q}} [U(\pi_s)] & > &  \mathbb{E}_{\vec{q}} [U(\pi_{pr})]\Rightarrow\\
r \hspace{5pt}  {-C^{R}_P(\pi_s)-C_E^R(\pi_s)} &>& ({-C^{R}_P(\pi_{pr})-C_{\tilde{G}}^R}-C_{\tilde{E}}^R(\pi_{pr}))\nonumber\\ &&\times (1-q_E-q_N)  \nonumber \\
&& +({-C^{R}_P(\pi_{pr})-C_E^R(\tilde{\pi}_{pr})-C_{\tilde{G}}^R}) \times q_E  \nonumber \\
&& +({-C^{R}_P(\pi_{pr})-C_E^R(\pi_{pr})}) \times q_N \nonumber
\end{eqnarray}%
}%
where $\mathbb{E}_{\vec{q}} [U(\pi)]$ denotes the expected utility of the robot under the human's observation policy (or mixed strategy) $\vec{q}$ if it chooses to make and execute the plan $\pi$. Note that the equation is linear w.r.t. the variables $q_N$ and $q_E$. Thus, there will be a region on one side of the linear boundary where the robot always executes $\pi_s$.\footnote{In repeated interaction settings when the stakes are high or the change in trust cannot be easily observed in a non-cooperative setting, our inference method for finding the trust boundary (when no pure Nash exists) still works while the increase/decrease of human’s trust can be modeled with the random variable that is a part of the game-theoretic model.}

\section{Experimental Setup and Evaluation} 

In this section, we first model a task-planning scenario in our game-theoretic framework. Then, we compute the proposed trust boundary, which provides an optimal monitoring strategy for the human, and leverage this in our human subject studies.

\subsection{Robot Delivery Domain}
Most motion planning scenarios only consider the execution phase (rather than modeling both the planning and execution stages separately), while task-planning domains concentrate only on the planning phase of the problem. Given that our game-theoretic model can account for both the stages, choosing an existing domain that renders itself naturally to both the planning and execution phases becomes a challenging task. To this extent, we choose the robot-delivery domain \cite{kulkarni2016explicable} because (1) we can use the task planning domain definition as-is, and (2) the domain has a straightforward interpretation for the execution stage.

This domain allows us to formulate realistic scenario to model the no-trust case with a human supervisor and a robot worker. The robot can collect parcels (that may not be waterproof) from the reception desk and/or coffee from the kitchen and deliver it to a particular location (eg. employee's desk). To do so, the robot has the following actions: \{{\em pickup, putdown, stack, unstack, move}\} which can be represented in the Planning Domain Definition Language (PDDL) \cite{kulkarni2016explicable}.

\subsubsection*{\textbf{Problem Instance}} The problem instance in our setting has the initial setting where (1) the robot is standing at a position equidistant to the reception and the kitchen, (2) there is a parcel located at the reception that is intended for the employee, (3) there is brewed coffee in the kitchen that needs to be delivered in a tray to the employee. The goal for the robot is to collect and deliver the coffee and the parcel to the employee.

\subsubsection*{\textbf{Robot Plans}}
In Figure \ref{fig:scenarios}, we show two plans in which the robot achieves the goal of collecting coffee from the kitchen and parcel from the reception desk and delivers them to an employees' desk. In the plan shown of the left $\pi_s$, the robot (1) collects coffee, (2) delivers it to the employee, (3) goes back along the long corridor to collect the parcel from the reception desk and finally (4) delivers it back to the same employee. In the plan on the right $\pi_{pr}$, the robot collects coffee from the kitchen, (2) collects parcel from the reception desk and puts them on the same tray and finally, (3) delivers both of them to the employee.\footnote{Given the (actual and the human's) domain models and the problem instance, these plans can simply be computed using available open-source software like {\tt Fast-Downward} or web-services like {\tt planning.domains}.}
\subsection{Computing the Trust Boundary in a Task-Planning Scenario}
In order to compute the trust boundary, we calculate the utility values for our game leveraging Table \ref{table:normal-game} and the cost incurred by $R$ and $H$ in this robot delivery domain.
As we have different types of costs for our game, we choose to normalize all of them to be $\in [0,1]$ and then used a multiplicative factor which represents the significance of each cost type.

In this example, if the robot makes $\pi_{pr}$, it will be executable (or safe) as per one of the two observers whose models make up the set $\mathcal{M}^{R}_{H}$. Thus, the robustness for $\pi_{pr}$ is $r = \frac{1}{2} = 0.5$. On the other hand, the plan $\pi_s$ is executable (and thus, overall safe) in both the models in $\mathcal{M}^{R}_{H}$.

\subsubsection*{\textbf{Robot Utility Values}}
We used the Fast Downward planner \cite{helmert2006fast} on the robot delivery domain \cite{kulkarni2016explicable} to find the execution costs for $R$. For $\pi_{pr}$ with $r=0.5$, it was $(C_E^R(\pi_{pr})=)10$ while for $\pi_s$, it was $(C_E^R(\pi_{s})=)14$.
We note that the time for coming up with the plan $\pi_s$ is $0.19$s whereas it is $0.177$s for coming up with $\pi_{pr}$ on a machine with an Intel Xeon CPU (clock speed 3.4 Ghz) and 128GB RAM. The unit for execution costs, although not well defined in PDDL models can be a stand in for the fuel costs used up by the robot while the planning costs is measured in seconds. Thus, we first normalize the planning cost and then choose an appropriate prioritization parameter to compare the planning and the execution costs. We obtain $C_P^R(\pi_{pr})=3.54$ and $C_P^R(\pi_s)=3.8$. Lastly, the penalty for not achieving the goal is a random variable with the Bernoulli distribution of $(1-r)$ where $C_{\tilde{G}}^R=\begin{cases}0 & r\\ 20 & 1-r \end{cases}$ which is double the size of the cost of execution in the non-zero case.

Given that the complexity of determining plan existence for classical planning problems is P-SPACE \cite{bylander1994computational}, a legitimate concern is how realistic is the idea of solving two planning problems to obtain the utility values for our game. To avoid this high computational cost, we can solve a relaxed version of these planning problems to obtain an approximation for the real plan cost. Note that this approximation in the utility space, only necessary for large instances, can result in sub-optimal monitoring strategies.

\begin{figure}[t]
    \centering
     \pgfmathdeclarefunction{gauss}{2}{%
      \pgfmathparse{1/(#2*sqrt(2*pi))*exp(-((x-#1)^2)/(2*#2^2))}%
    }
    
    \qquad
    \begin{adjustbox}{width=0.27\textwidth}
        \centering
        \begin{tikzpicture}[
        every node/.style={text=black},
        fdesc/.style={anchor=south east,sloped,font=\scriptsize,pos=#1},
        fdesc/.default=1,
      ]
      \begin{axis}[
          height=4.5cm, width=4.5cm,
          ymin=-0.2,
          ymax=1,
          xmin=-0.2,
          xmax=1,
          xtick={0,1},
          ytick={0,1},
          tick label style={font=\tiny},
          axis lines*=left, xlabel=$q_N$, ylabel=$q_E$,
          every axis y label/.style={at=(current axis.above origin),anchor=south east},
          every axis x label/.style={at=(current axis.right of origin),anchor=north west},
       xlabel style={font=\tiny},
           ylabel style={font=\tiny},
          enlarge y limits=upper,
          enlarge x limits,
          samples=2 
        ]
        \addplot[Maroon,domain=-0.25:1,name path=p15] {(-5.74+10*x)/(3)};
        \addplot[gray,domain=0:1,name path=addone] {1-x} node[fdesc=0.65]{\tiny $q_E + q_N = 1$}node[below right]{};
        \addplot[gray,domain=0:1,name path=y0] {0};
        \addplot[gray,domain=0:6,name path=x0] coordinates {(0,0)(0,1)};
       
        \addplot[pattern=horizontal lines, pattern color=Maroon!60] fill between[of=p15 and addone, soft clip={domain=.565:0.672}];
        \addplot[pattern=horizontal lines, pattern color=Maroon!60] fill between[of=y0 and addone, soft clip={domain=0:0.565}];
        \addplot[fill=red!80!black] (0,0) circle (1.5pt) node[below]{\color{red!80!black}{\tiny ab~~~~Explicable plans \cite{kulkarni2016explicable}}};
        \addplot[fill=red!80!black] (0,1) circle (1.5pt) node[above]{\color{red!80!black}{\tiny Legible plans \cite{dragan2013legibility}}};
      \end{axis}
    
        \end{tikzpicture}
    \end{adjustbox}
    \caption{An observation strategy in the trust region (shaded) ensures that the robot sticks to $\pi_s$. This shows one can reduce monitoring costs while ensuring explicable/legible/safe behavior.}
    \label{fig1}
\end{figure}

\subsubsection*{\textbf{Human Utility Values}}
We have two possible supervisors with two different mental models. In one, the second plan $\pi_{pr}$ is unsafe because the coffee and parcel taken in the same tray runs the risk of the spilling coffee and ruining the package. In the other, both plans are considered safe. Lastly, note that the length of the corridor is a key factor in determining how sub-optimal $\pi_s$ is for the robot to execute when compared to $\pi_{pr}$ because, for $\pi_s$, the robot requires an extra trip back to the reception (i.e. two extra traversals of the corridor).

We consider the cost for the human to observe the plan to be proportional to the planning time for $R$ because the plans that took a longer time to be built will need $H$ to spend a longer time to reason about it correctness and/or optimality. Thus, $C_P^H(\pi_{pr})=0.885$ and $C_P^H(\pi_s)=0.95$.
The cost incurred by the human when they observe the execution of plan $\pi_s$ is $8$ while $C_E^H(\pi_{pr})=4$ assuming that the cost of going through the long corridor is $2$ (note that the difference in observation cost increases as this value increases). However, if the human thinks carrying the parcel and the coffee in a single tray is unsafe, the cost of the observation of the partial execution of the plan is $1.5$ because it will stop the robot as soon as it tries to put them on the same tray.
For the inconvenience costs, we have the Bernoulli distribution in which the non-zero case is the same as the cost of observation for the safe plan, since if the robot does something unsafe the human have to stop it and make it to do the safe plan. So, we have
\[I_{P}^H=\begin{cases}0 & r\\ 0.95 & 1-r \end{cases} \text{and } I_E^H=\begin{cases}0 & r\\ 8 & 1-r \end{cases}
\]%
The cost $V_{I}^{H}$'s can be calculated as the model difference between the least and most constrained models in $\mathcal{M}^{R}_{H}$ in terms of the number of preconditions and effects of actions. Lastly, if an unsafe plan runs to completion, the overall magnitude of this variable is higher. After calculation, $V_{I}^H=\begin{cases}0 & r\\ 20 & 1-r \end{cases}$.\\

We can now define the utility matrix for the players ${(R, {\color{MidnightBlue} H})}$ as follows, \\
First type with probability $0.5$:
\begin{eqnarray}
\label{eq5}
{ \left[ {\begin{array}{ccc}
   (-13.54, {\color{MidnightBlue} -0.885}) & (-13.54, {\color{MidnightBlue} -4}) & (-13.54, {\color{MidnightBlue}0}) \\ (-17.80, {\color{MidnightBlue}-0.95})  & (-17.80,  {\color{MidnightBlue}-8.00}) & (-17.80,  {\color{MidnightBlue} 0}) 
  \end{array} } \right] } \nonumber 
\end{eqnarray}%
Second type with probability $0.5$:
\begin{eqnarray}
\label{eq6}
{ \left[ {\begin{array}{ccc}
   (-23.54, {\color{MidnightBlue} -1.835}) & (-26.54, {\color{MidnightBlue} -9.5}) & (-13.54, {\color{MidnightBlue}-20}) \\ (-17.80, {\color{MidnightBlue}-0.95})  & (-17.80,  {\color{MidnightBlue}-8.00}) & (-17.80,  {\color{MidnightBlue} 0}) 
  \end{array} } \right] } \nonumber 
\end{eqnarray}%
\subsection{Trust Boundary Calculation}
According to Proposition 1, this game does not have a pure Nash Eq. strategy with probability $0.5$. Therefore, we now find the boundary in the space of mixed strategies for second type of $H$ who can choose to adopt which will ensure that the robot always executes $\pi_s$.
To do so, we use the values defined above and plug them into equation \ref{eqn:not} and obtain,
\begin{eqnarray}
\label{eq7}
\qquad ~~10\times q_N -3\times q_E -5.74 & < & 0
\end{eqnarray}%
In Figure \ref{fig1}, we plot the trust boundary represented by the lines in Eqn. \ref{eq7}. 
The three black lines (sides of the larger triangle) represent the feasible region for the human's mixed strategy $\vec{q}$. Monitoring strategy in the shaded region guarantees the robot, being a rational agent, executes $\pi_{s}$. The strategy that optimizes $H$'s monitoring cost and yet ensures the robot adheres to $\pi_s$ lies on the trust boundary indicated using the red line. Note that existing work in task \cite{kulkarni2016explicable} and motion \cite{dragan2013legibility} planning that ensures explicable and legible behavior expects pure strategies for observing the plan and observing the execution respectively. This restricts the humans to only two corners of the feasible strategy space, hardly optimizing the human's cost.

\begin{figure}[t]
\includegraphics[width=0.45\textwidth,trim={0 0 0 3.5cm},clip]{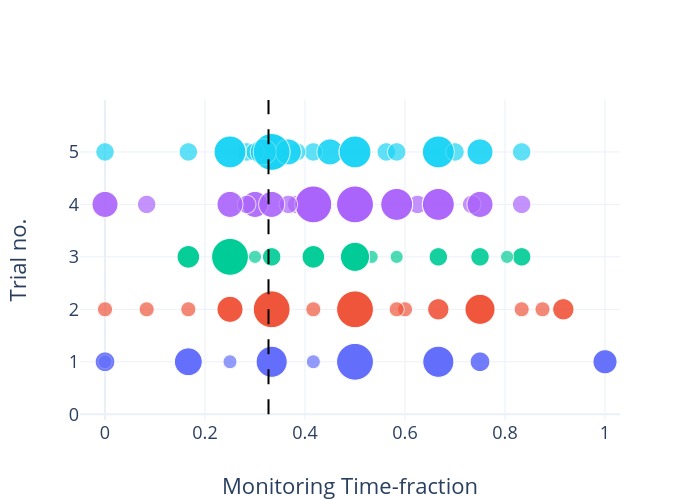}
\caption{Participant's monitoring strategies across multiple trials. Trust boundary indicated using the black vertical line.}
\label{fig:hs}
\end{figure}
\subsection{Human Studies}
We conduct two human-subject studies. In the first study, we seek to ascertain the necessity of our contribution to model the interaction in a game-theoretic formulation that computes an optimal monitoring strategy (eg. humans may simply be able to figure out a good strategy by just performing the monitoring task by themselves). Given the results of the first study establish grounds for a better approach, we evaluate how effective our method is at helping human participants optimize their monitoring strategy. Specifically, our studies seek to validate three hypotheses:
\begin{itemize}
    \item[\textbf{H1:}] The inherent monitoring strategies adopted by human are going to be inferior to the optimal monitoring strategy (that incurs lower monitoring cost while ensuring safe robot behavior)
    \item[\textbf{H2:}] Humans tend to deviate from always monitoring the robot (doing which can lead the robot to choose unsafe behaviors)
    \item[\textbf{H3:}] If the optimal monitoring strategy computed by our game-theoretic formulation is provided to humans, they will \textit{follow} it and it \textit{helps} them to come up with better monitoring strategy.
\end{itemize}

Note that \textbf{H2} contradicts the inherent assumption made in earlier work \cite{zhang2017plan,dragan2013legibility}, at least in the context of the robot-supervision scenario.  Our first study seeks to validate \textbf{H1} and \textbf{H2} while the second study validates \textbf{H1} and \textbf{H3}.

\subsubsection{\textbf{Study I: Do we need this service?}}
Participants in this study play the role of a student in a robotics department who are asked to monitor the robot for an hour. To make the monitoring action be associated with a cost, we consider a second task where participants can choose to grade exam papers (and get paid) instead of monitoring the robot. Given the scarcity of participants who have experience as a professional supervisor, we combine the actions to monitor the plan and monitor the execution as a single `monitor the robot' action to simplify the scenario. The combination of the planning and execution phase simply helps to reduce the human’s action set; helping them easily understand the setting and choose between a fewer number of actions. The other action `grade exam papers' represents the action to not-monitor the robot. As opposed to asking the participants for mixed strategies over the two actions, which is hard for them to interpret, we ask them to give us a time slice for which they would choose a particular action (eg. 30 minutes to monitor the robot and 30 minutes to grade exam papers). We provide the participants with their utility values for their actions conditioned on the robot's pure strategies (i.e. the plans $\pi_{s}$ and $\pi_{pr}$). We inform them that the robot may have incentive to consider a less costly (but probably risky) plan depending on the fraction of time allocated for monitoring. We let each participant do five trials and after each trial, the overall utility based on the participant's monitoring strategy and the robot's strategy is reported to them. The robot does not adapt itself to the human's strategy in the previous trial (which intends to preserve the non-repeated nature of our game).

\begin{figure}[t]
\includegraphics[width=0.45\textwidth,trim={0 0 0 3.5cm},clip]{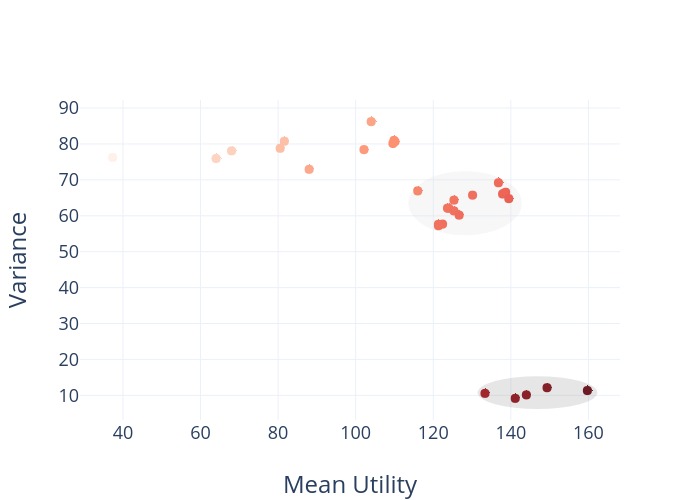}
\caption{Average utility and variance for each participant across the five trials.}
\label{fig:hs_ind}
\end{figure}

A pilot study was first run on $4$ participants whose feedback helped us fix several issues in the interface that inhibited clarity. We then collected data by asking $32$ participants to undertake the study. The participants of this study were all graduate students across various engineering departments at our university. 


{\em Aggregate Results -- Changes in Monitoring Strategy across Trials: }
Note that a participant, given the information on the interface, can formulate a simplified version of the game-theoretic model proposed in this paper and find the optimal strategy for monitoring (which is to monitor the robot for $0.327$ or $19.62$ minutes of an hour and use the remaining time to grade papers). The participants' time slice allocated for monitoring, across the five trials, are shown in Fig. \ref{fig:hs}. Given that there are only two actions for the participant, the strategy can be represented using a single variable (fraction to monitor the robot) and thus, is plotted along the x-axis. The size of each bubble is proportional to the number of participants who selected a particular strategy. The optimal strategy is shown using a black vertical line (i.e. $x=0.327$). In the first trial, we noticed a small subset of users ($n=5$) calculate the (almost) optimal strategy using the utility values specified on the interface. Majority of the other users ($n=18$) choose a risk-averse strategy, i.e. monitor the robot to ensure it performs a safe plan even if it meant losing out on money that could be earned from grading. The remaining $9$ participants, in the hope of making more money, spent a larger time grading papers but, eventually ended up with a lower reward because the robot performed the risky plan that failed to achieve the goal.

We observed that participants discarded extreme strategies (i.e. only monitor or only grade papers) in later trials and started considering strategies that strike a better balance. This only seems natural given that we provided them feedback after each trial. We believe that the feedback helped the participants improve their strategies via trial-and-error; note that they did not consider using the provided utility values to come up with a near-optimal strategy. In Fig \ref{fig:hs}, note that for the first two trials, the strategies are well spread out in the range $[0,1]$ where as in the last two trials, the strategies are clustered more densely, with few data points below $0.25$ or above $0.75$. Even then, we observe that in the final trail, the difference in the distribution of the human-selected strategy and the optimal strategy is statistically significant (with a p-value of $0.0052$). This conclusion supports \textbf{H1}, demonstrating that humans cannot come up with an optimal monitoring strategy on their own. At best, they learn to avoid certain strategies via repeated trial-and-error (which may not always be possible in the real-world). 

{\em Participant Types: }
In Figure \ref{fig:hs_ind}, we plot the average utility of each participant across five trials on the x-axis. The y-axis represents the variance. Highlighted in dark, at the bottom right, are five participants that chose observation probabilities in the trust region but not exactly on the trust boundary, i.e. sub-optimal w.r.t. the optimal monitoring strategy that yields a reward of $173.77$. Although these five participants defaulted to a greedy behavior (that reduced the observation time and made more money by grading papers) after the first trial, they explored cautiously-- only deviating slightly from the good policies they initially discovered.
Towards the top-right corner, the set of points circled in light gray, we see a dense cluster of participants ($=15$) who obtained a high average utility but tried to tweak their strategies significantly-- they monitored less, allowing the robot to choose the riskier plan that lead to a large loss. This implied that humans deviate to more split-time strategies and error on the side of monitoring less (i.e land on the unsafe region of the trust boundary) (\textbf{H2}).



{\em Subjective Evaluation:} We asked each participant two subjective questions-- (1) how did they come up with a particular monitoring strategy and (2) would they consider an algorithm that suggests them an optimal strategy. Out of the $30$ participants who answered (1), most of them identified the tension that exists between choosing a relaxed monitoring strategy and the robot considering unsafe behaviors. Of these, $12$ participants identified the scenario as an optimization problem; others resorted to trial-and-error.\\
For question (2), $24$ out of the $31$ candidates answered {\em yes}. On being asked why (as a follow-up), they all expected the software would be (1) faster and (2) maximize their utility. Three participants said they were willing to use it if it was just a suggestion, while one participant felt they would only need it for large scale problems. A participant said that they would place their trust on the software only if they knew that the developer had a strong background in mathematics. This inclination to use a software sets the stage aptly for our next study. 

\subsubsection{\textbf{Study II: Does this service help?}}
In this study, we designed a user interface that simulates the robot delivery domain where the participant has to monitor the robot. Similar to the previous setting, we consider a second task of labeling images that earns extra points (and an additional payment). We convert the whole robot task execution to designated steps (e.g. $29$ steps for executing $\pi_s$). Figure 5 depicts the map that is shown to the participants. Each participant has $7$ rounds to monitor the robot task execution step-by-step. Note that the analysis we undertake provides the human an additional advantage absent from the single-shot interaction setting we are primarily interested in. By allowing for data collected from the same participant over multiple interactions, we are in principle allowing the human the possibility of coming up with more informed monitoring strategies, a possibility absent in the original single-shot setting. At any step, they can choose to stop monitoring the robot and move on to the image labeling task. The participants' utility values are represented as points and shown in the table \ref{table:points}. We also informed participants that the robot adjusts its behavior based on their monitoring time. So, if they monitor the robot long enough, the robot will do safe behavior; otherwise, it can execute risky behavior in the current and the next round.
\begin{figure}[t]
\label{fig:map_main}
\begin{subfigure}[\label{fig:risky}]
    \centering
    \includegraphics[width=.235\textwidth]{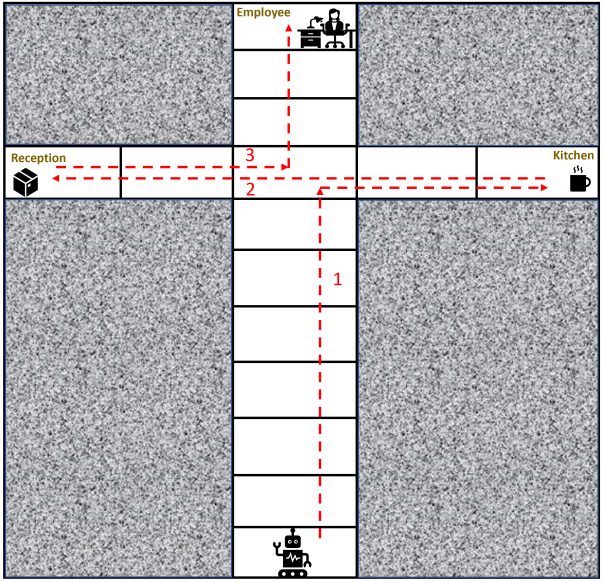}
    \end{subfigure}
    \begin{subfigure}[\label{fig:safe}]
    \centering
    \includegraphics[width=.235\textwidth]{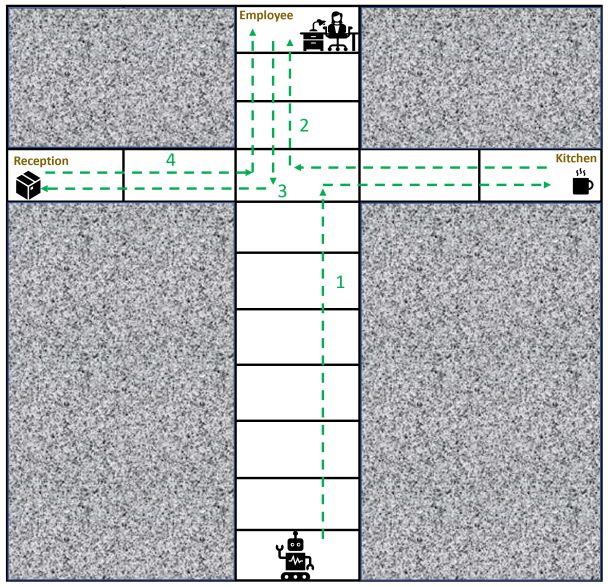}
    \label{fig:description}   
    \end{subfigure}
    \caption{The map that is shown to the participants. Given the human monitoring strategy, the robot either will execute the safe plan $\pi_s$ or the probably risky plan $\pi_{pr}$ (a) The probably risky plan (22 steps), (b) The safe plan (29 steps). Each move on the map (e.g. moving through each block, picking up the objects) is considered a step of the plan execution}
\end{figure}
We recruited a total of $26$ participants (students at our university) for this study. Each subject was paid $\$5$ for participation and for every $100$ points earned, they can received an additional $60$ cents. Negative points did not reduce the base payment.\\
We considered two conditions-- (1) \textbf{Treatment Case}: we suggest the optimal strategy to, and (2) \textbf{Control Case} we don't provide the optimal strategy (similar to the previous setting). In our between-subject evaluation, we divided our participants into two equal halves for each condition (see {\em Supplementary Material} for details).

Based on our algorithm, we computed trust boundary (see Equation \ref{eq7}), and the optimal strategy is $x=0.327$. As we converted the whole monitoring time to steps of monitoring, the computed optimal strategy is to monitor the robot for $10$ steps (and monitoring for $>10$ steps encodes the trust region). In the \textbf{\em Treatment Case} the participants were told the minimum number of steps they need to monitor the robot (i.e. 10) to ensure safe behavior. We further specified that this was a recommendation that they may or may-not choose to follow.
In the \textbf{\em Control Case}, everything was kept the same except that no recommendation was given.

\begin{table}[tp]
\centering
\caption{Summary table of costs}
\label{table:points}
\resizebox{\columnwidth}{!}{%
\begin{tabular}{p{8cm}c}
\toprule
       Description & Points \\
\midrule
Monitoring $R$ the whole time; $R$ does $\pi_s$ & $0$   \\
Labeling images the whole time & $+200$  \\
Not monitor $R$ enough; $R$ does $\pi_{pr}$ & $-200$\\
Monitoring $R$ enough, $R$ does $\pi_{pr}$ (because of not monitoring it enough in previous round) & $-40$\\
Point ratio for image labeling & $(\frac{\text{\# steps left}}{\text{total \# steps}})* 200$\\
\bottomrule
\end{tabular}}
\end{table}
\begin{figure}[t]
\includegraphics[width=0.5\textwidth]{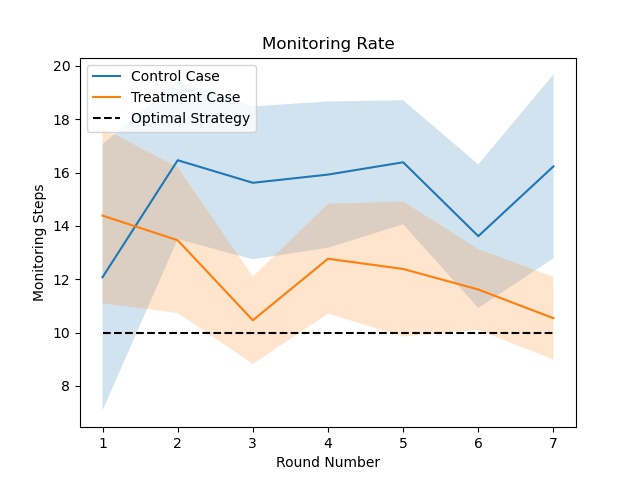}
\caption{Mean and std-dev. of steps monitored in each round.}
\label{fig:out}
\end{figure}

\subsubsection*{\textbf{Results}}
Across the two conditions, we collected the number of monitoring steps selected by a participant in each round (see \autoref{fig:out}). Participants in the treatment case followed the optimal strategy or selected strategies closer to the optimal strategy compared to participants in the control case.

By performing a one-tailed p-value test via t-test for independent means, we were able to validate \textbf{H1} and \textbf{H3} with results being significant at p-value of $<0.05$. First, we compared mean over all rounds to compare the treatment and the control case. A p-value of $0.0145$ shows that the participants monitor the robot differently in the two conditions. This coupled with the fact that participants had near-optimal strategy in the treatment case validates \textbf{H3}.
Further, we tested if the strategy in the final round is different from the optimal strategy. For the control case, we observed a p-value of $0.004$ showing statistical significance, whereas we observed a p-value of $0.279$ for the treatment case, demonstrating there was no statistically significant different between the human's strategy and the optimal one. This result reinforces that \textbf{H1} holds-- human cannot discover the optimal monitoring strategy by themselves. Further, it also shows that our framework can effectively assist in humans developing more optimal strategy (reinforcing \textbf{H3} holds).

\section{Conclusions and Future Work}

We model the notion of trust that a human supervisor places on a worker robot by modeling this interaction as a Bayesian Game. The particular Human-Robot interaction setting situates our work at the middle of the spectrum that ranges from fully-cooperative settings on one end to fully-adversarial scenarios on the other. We show that existing notions of game-theoretic trust break down in our setting when the worker robot cannot be trusted due to the absence of pure strategy Nash Equilibrium. Thus, we introduce a notion of trust boundary that optimizes the supervisor's monitoring cost while ensuring that the robot workers stick to safe plans.
Given that supervisors or caretakers often spend time working on side goals (such as talking over the phone, sleeping, watching movies, etc.), we carefully design a human study to see whether humans have an inherent sense of good monitoring policies. Beyond objective results, we show that most humans explicitly say that they would prefer an algorithm that computes the optimal strategy for them (in our case, located on an edge of the trust region). 
Such strategies can also be useful in other scenarios where the supervised agent is not a robot. Note that in those cases, the formulation needs to capture the irrationality and computational capabilities of the monitored agent. In another human subject study, we evaluated whether the human will follow the given optimal strategy and showed that our framework can indeed assist the human to follow a better monitoring strategy.

In future, we plan to study the notion of trust that is prevalent in repeated interaction settings. An interesting problem that may occur in such settings is when the robot primes the human to not observe its behavior by choosing safe and sub-optimal behaviors (thereby engendering trust) only to exploit it in a high-stake scenario.

\section*{Acknowledgments} This research is supported in part by ONR grants N00014-16-1-2892, N00014-18-1- 2442, N00014-18-1-2840, N00014-9-1-2119, AFOSR grant FA9550-18-1-0067, DARPA SAIL-ON grant W911NF-19- 2-0006, and a JP Morgan AI Faculty Research grant. We thank Sarath Sreedharan for his helpful feedback and comments on the second user study.


\bibliographystyle{IEEEtran}
\bibliography{main_THMS.bib}

\begin{thebibliography}{10}
\providecommand{\url}[1]{#1}
\csname url@samestyle\endcsname
\providecommand{\newblock}{\relax}
\providecommand{\bibinfo}[2]{#2}
\providecommand{\BIBentrySTDinterwordspacing}{\spaceskip=0pt\relax}
\providecommand{\BIBentryALTinterwordstretchfactor}{4}
\providecommand{\BIBentryALTinterwordspacing}{\spaceskip=\fontdimen2\font plus
\BIBentryALTinterwordstretchfactor\fontdimen3\font minus
  \fontdimen4\font\relax}
\providecommand{\BIBforeignlanguage}[2]{{%
\expandafter\ifx\csname l@#1\endcsname\relax
\typeout{** WARNING: IEEEtran.bst: No hyphenation pattern has been}%
\typeout{** loaded for the language `#1'. Using the pattern for}%
\typeout{** the default language instead.}%
\else
\language=\csname l@#1\endcsname
\fi
#2}}
\providecommand{\BIBdecl}{\relax}
\BIBdecl

\bibitem{chen2018planning}
M.~Chen, S.~Nikolaidis, H.~Soh, D.~Hsu, and S.~Srinivasa, ``Planning with trust
  for human-robot collaboration,'' in \emph{Proceedings of the 2018 ACM/IEEE
  International Conference on Human-Robot Interaction}, 2018, pp. 307--315.

\bibitem{xu2015optimo}
A.~Xu and G.~Dudek, ``Optimo: Online probabilistic trust inference model for
  asymmetric human-robot collaborations,'' in \emph{2015 10th ACM/IEEE
  International Conference on Human-Robot Interaction (HRI)}.\hskip 1em plus
  0.5em minus 0.4em\relax IEEE, 2015, pp. 221--228.

\bibitem{jacovi2021formalizing}
A.~Jacovi, A.~Marasovi{\'c}, T.~Miller, and Y.~Goldberg, ``Formalizing trust in
  artificial intelligence: Prerequisites, causes and goals of human trust in
  ai,'' in \emph{Proceedings of the 2021 ACM conference on fairness,
  accountability, and transparency}, 2021, pp. 624--635.

\bibitem{goodrich2003seven}
M.~A. Goodrich and D.~R. Olsen, ``Seven principles of efficient human robot
  interaction,'' in \emph{SMC'03 Conference Proceedings. 2003 IEEE
  International Conference on Systems, Man and Cybernetics. Conference
  Theme-System Security and Assurance (Cat. No. 03CH37483)}, vol.~4.\hskip 1em
  plus 0.5em minus 0.4em\relax IEEE, 2003, pp. 3942--3948.

\bibitem{sankaranarayanan2007towards}
V.~Sankaranarayanan, M.~Chandrasekaran, and S.~Upadhyaya, ``Towards modeling
  trust based decisions: a game theoretic approach,'' in \emph{European
  Symposium on Research in Computer Security}.\hskip 1em plus 0.5em minus
  0.4em\relax Springer, 2007, pp. 485--500.

\bibitem{kulkarni2016explicable}
A.~Kulkarni, T.~Chakraborti, Y.~Zha, S.~G. Vadlamudi, Y.~Zhang, and
  S.~Kambhampati, ``Explicable robot planning as minimizing distance from
  expected behavior,'' \emph{CoRR, abs/1611.05497}, 2016.

\bibitem{dragan2013legibility}
A.~D. Dragan, K.~C. Lee, and S.~S. Srinivasa, ``Legibility and predictability
  of robot motion,'' in \emph{Proceedings of the 8th ACM/IEEE international
  conference on Human-robot interaction}.\hskip 1em plus 0.5em minus
  0.4em\relax IEEE Press, 2013, pp. 301--308.

\bibitem{zhang2017plan}
Y.~Zhang, S.~Sreedharan, A.~Kulkarni, T.~Chakraborti, H.~H. Zhuo, and
  S.~Kambhampati, ``Plan explicability and predictability for robot task
  planning,'' in \emph{Robotics and Automation (ICRA), 2017 IEEE International
  Conference on}.\hskip 1em plus 0.5em minus 0.4em\relax IEEE, 2017, pp.
  1313--1320.

\bibitem{kockemann2014grandpa}
U.~K{\"o}ckemann, F.~Pecora, and L.~Karlsson, ``Grandpa hates
  robots-interaction constraints for planning in inhabited environments.'' in
  \emph{AAAI}, 2014, pp. 2293--2299.

\bibitem{hadfield2016off}
D.~Hadfield{-}Menell, A.~D. Dragan, P.~Abbeel, and S.~J. Russell, ``The
  off-switch game,'' in \emph{{IJCAI}}.\hskip 1em plus 0.5em minus 0.4em\relax
  ijcai.org, 2017, pp. 220--227.

\bibitem{sreedharan2017explanations}
S.~Sreedharan, S.~Kambhampati \emph{et~al.}, ``Explanations as model
  reconciliation—a multi-agent perspective,'' in \emph{2017 AAAI Fall
  Symposium Series}, 2017.

\bibitem{johnson2020impact}
E.~Johnson and J.~Gratch, ``The impact of implicit information exchange in
  human-agent negotiations,'' in \emph{Proceedings of the 20th ACM
  International Conference on Intelligent Virtual Agents}, 2020, pp. 1--8.

\bibitem{kim2017collaborative}
J.~Kim, C.~Banks, and J.~Shah, ``Collaborative planning with encoding of users'
  high-level strategies,'' in \emph{Proceedings of the AAAI Conference on
  Artificial Intelligence}, vol.~31, no.~1, 2017.

\bibitem{paruchuri2008playing}
P.~Paruchuri, J.~P. Pearce, J.~Marecki, M.~Tambe, F.~Ordonez, and S.~Kraus,
  ``Playing games for security: An efficient exact algorithm for solving
  bayesian stackelberg games,'' in \emph{Proceedings of the 7th international
  joint conference on Autonomous agents and multiagent systems-Volume 2}.\hskip
  1em plus 0.5em minus 0.4em\relax International Foundation for Autonomous
  Agents and Multiagent Systems, 2008, pp. 895--902.

\bibitem{sinha2015physical}
A.~Sinha, T.~H. Nguyen, D.~Kar, M.~Brown, M.~Tambe, and A.~X. Jiang, ``From
  physical security to cybersecurity,'' in \emph{Journal of Cybersecurity},
  vol.~1, no.~1.\hskip 1em plus 0.5em minus 0.4em\relax Oxford University
  Press, 2015, pp. 19--35.

\bibitem{sengupta2017game}
S.~Sengupta, S.~G. Vadlamudi, S.~Kambhampati, A.~Doup{\'e}, Z.~Zhao,
  M.~Taguinod, and G.-J. Ahn, ``A game theoretic approach to strategy
  generation for moving target defense in web applications,'' in
  \emph{Proceedings of the 16th Conference on Autonomous Agents and MultiAgent
  Systems}.\hskip 1em plus 0.5em minus 0.4em\relax International Foundation for
  Autonomous Agents and Multiagent Systems, 2017, pp. 178--186.

\bibitem{schlenker2018deceiving}
A.~Schlenker, O.~Thakoor, H.~Xu, F.~Fang, M.~Tambe, L.~Tran-Thanh, P.~Vayanos,
  and Y.~Vorobeychik, ``Deceiving cyber adversaries: A game theoretic
  approach,'' in \emph{Proceedings of the 17th International Conference on
  Autonomous Agents and MultiAgent Systems}.\hskip 1em plus 0.5em minus
  0.4em\relax International Foundation for Autonomous Agents and Multiagent
  Systems, 2018, pp. 892--900.

\bibitem{nguyen2017robust}
T.~Nguyen, S.~Sreedharan, and S.~Kambhampati, ``Robust planning with incomplete
  domain models,'' \emph{Artificial Intelligence}, vol. 245, pp. 134--161,
  2017.

\bibitem{helmert2006fast}
M.~Helmert, ``The fast downward planning system,'' \emph{Journal of Artificial
  Intelligence Research}, vol.~26, pp. 191--246, 2006.

\bibitem{bylander1994computational}
T.~Bylander, ``The computational complexity of propositional strips planning,''
  \emph{Artificial Intelligence}, vol.~69, no. 1-2, pp. 165--204, 1994.

\end{thebibliography}

\end{document}